\documentclass[10pt,twocolumn,letterpaper]{article}

\usepackage{cvpr}
\usepackage{times}
\usepackage{epsfig}
\usepackage{graphicx}
\usepackage{amsmath}
\usepackage{amssymb}
\usepackage{balance}
\usepackage{algorithm}
\usepackage{bbm}
\usepackage{algpseudocode}
\usepackage[pagebackref=true,breaklinks=true,letterpaper=true,colorlinks,bookmarks=false]{hyperref}

\usepackage{booktabs,multirow,rotating}

\def\ie{\emph{i.e}\onedot}

\def\wrt{w.r.t\onedot} 
\def\etal{\emph{et al}\onedot}
\makeatother

\def\Vec#1{{\boldsymbol{#1}}}

\newcommand{\eat}[1]{}

\usepackage[breaklinks=true,bookmarks=false]{hyperref}

\cvprfinalcopy 

\def\cvprPaperID{1227} 

 \ifcvprfinal\fi
\begin{document}
	
	\title{Dense Regression Network for Video Grounding}
	
	\author{
		Runhao~Zeng$^{1,3}$\thanks{This work was done when Runhao Zeng was a research intern at Peng Cheng Laboratory, Shenzhen, China.}~~~~Haoming Xu$^{1}$~~~~Wenbing~Huang$^{4}$~~~~Peihao~Chen$^{1}$~~~~Mingkui~Tan$^{1}$\thanks{Corresponding author}~~~~Chuang~Gan$^{2}$\\
		$^{1}$School of Software Engineering, South China University of Technology, China \\ 
		$^{2}$MIT-IBM Watson AI Lab~~~~$^{3}$Peng Cheng Laboratory, Shenzhen, China \\
		$^{4}$Beijing National Research Center for Information Science and Technology (BNRist), \\
        Department of Computer Science and Technology, Tsinghua University\\
		{\tt\small \{runhaozeng.cs, ganchuang1990\}@gmail.com, hwenbing@126.com,
			mingkuitan@scut.edu.cn}
	}


	\maketitle
	\pagestyle{empty}
	\thispagestyle{empty}

\begin{abstract}
We address the problem of video grounding from natural language queries. The key challenge in this task is that one training video might only contain a few annotated starting/ending frames that can be used as positive examples for model training. Most conventional approaches directly train a binary classifier using such imbalance data, thus achieving inferior results. The key idea of this paper is to use the distances between the frame within the ground truth and the starting (ending) frame as dense supervisions to improve the video grounding accuracy.  Specifically, we design a novel dense regression network (DRN) to regress the distances from each frame to the starting (ending) frame of the video segment described by the query.  We also propose a simple but effective IoU regression head module to explicitly consider the localization quality of the grounding results (i.e., the IoU between the predicted location and the ground truth). Experimental results show that our approach significantly outperforms state-of-the-arts on three datasets (i.e., Charades-STA, ActivityNet-Captions, and TACoS). 
\end{abstract}

	\section{Introduction}
	
\begin{figure}[!t]
		\centering
		\includegraphics[width=\linewidth]{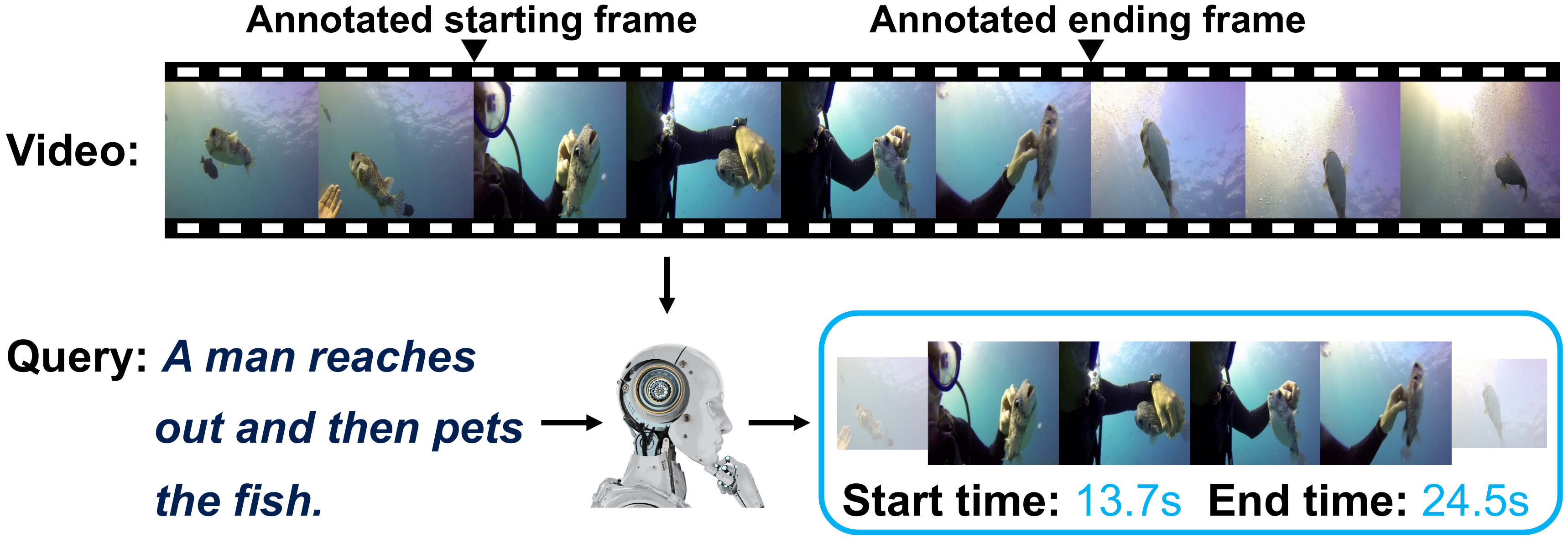}
		\caption{An illustrative example of the video grounding task. Given a video and a query, the video grounding task aims to identify the starting and ending time of the video segment described by the query. One key challenge  of this task is how to leverage dense supervision upon sparsely annotated starting and ending frames.}
		\label{fig:Brief}
		\vspace{-0.15in}
	\end{figure}

	Video grounding is an important yet challenging task in computer vision, which requires the machine to watch a video and localize the starting and ending time of the target video segment that corresponds to the given query, as shown in Figure~\ref{fig:Brief}. This task has drawn increasing attention over the past few years due to its vast potential applications in video understanding ~\cite{wang2016temporal, carreira2017quo,zeng2019graph,zeng2019breaking,chen2019relation}, video retrieval ~\cite{yu2018a, dong2019dual}, and human-computer interaction~\cite{singha2018dynamic,huang2020graph,zhu2019auxrn}, etc.
   

 The task, however, is very challenging due to several reasons: 1) It is nontrivial to build connections between the query and complex video contents; 2) Localizing actions of interest precisely in a video with complex backgrounds is very difficult. More critically, a video can often contain many thousands of frames, but it  may have only a few annotated starting/ending frames (namely the positive training examples), making the problem even more challenging.
    Previous approaches often adopt a two-stage pipeline~\cite{gao2017tall, xu2019multilevel, ge2019mac}, where they generate the proposals and rank them according to their similarities with the query. However, this pipeline incurs two issues: 1) One video often contains thousands of proposals, resulting in a heavy computation cost when comparing proposal-query pairs.
	2) The performance highly relies on the quality of proposals. 
	To address the above issues, one-stage video grounding methods~\cite{chen2019localizing,yuan2019to,ghosh2019excl} have been studied. Yuan~\etal~\cite{yuan2019to} propose to learn a representation of the video-query pair and use a multi-layer perceptron (MLP) to regress the starting and ending time. 
	Chen~\etal~\cite{chen2019localizing} and Ghosh~\etal~\cite{ghosh2019excl} attempt to predict two probabilities at each frame, which indicate whether this frame is a starting (or ending) frame of the target video segment. 
	The grounding result is obtained by selecting the frame with the largest starting (or ending) probability.
	However, the existing two-stage and one-stage methods have one common issue: they neglect the rich information from the frames within the ground truth.
	
	Recently, anchor-free approaches~\cite{huang2015densebox, redmon2016you, tian2019fcos, law2018cornernet, kong2019foveabox} for one-stage object detection become increasingly popular because of their simplicity and effectiveness. In this vein, Tian~\etal~\cite{tian2019fcos} propose the FCOS framework to solve object detection in a per-pixel prediction fashion. Specifically, FCOS trains a regression network to directly predict the distance from each pixel in the object to the object's boundary. 
	This idea is helpful for video grounding. If we train a model to predict the distance from each frame to the ground truth boundary, then all the frames within the ground truth can be leveraged as positive training samples. In this way, the number of positive samples is sufficiently increased and thus benefits the training. 
	
	
	
	In this paper, we propose a dense regression network for video grounding, which consists of four modules, including a video-query interaction module, a location regression head, a semantic matching head, and an IoU regression head.
	The main idea is as straightforward as training a regression module to directly regress the ground truth boundary from each frame within the ground truth.
	In the training, all frames within the ground truth are selected as positive samples. 
	By doing so, the sparse annotation is able to be used to generate more positive training samples sufficiently, which boosts grounding performance eventually.
	
	For each video-query pair, our model produces dense predictions (\ie, one predicted temporal bounding box for each frame) while we are only interested in the one that matches the query best. To select the best grounding result, we focus on two perspectives: 1) Does the box match the query semantically? 2) Does the box match the temporal boundary of the ground truth? Specifically, we train a semantic matching head to predict a score for each box, which indicates whether the content in the box matches the query semantically. However, this score cannot directly reflect the localization quality (\ie, the IoU with the ground truth), which is of vital importance for video grounding. This motivates us to further consider the localization quality of each prediction. To do so, one may use the ``centerness" assumption in FCOS, which, however, is empirically found inapplicable for video grounding (see Table~\ref{tab:iou}).
	In this paper, we train an IoU regression head to directly estimate the IoU between the predicted box and the ground truth. Last, we combine the matching score and the IoU score to find the best grounding result.
	It is worth noting that the dense regression network works in a one-stage manner.
	We evaluate our proposed method on three popular benchmarks for video grounding, \ie, Charades-STA~\cite{gao2017tall}, ActivityNet-Captions~\cite{krishna2017dense} and TACoS ~\cite{regneri2013grounding}.
	
	
	To sum up, our contributions are as follows:
	\begin{itemize}
		\item We propose a dense regression network for one-stage video grounding. We provide a new perspective to leverage dense supervision from the sparse annotations in video grounding.
		
		\item To explicitly consider the localization quality of the predictions, we propose a simple but effective IoU regression head and integrate it into our one-stage paradigm. 
		
		\item We verified the effectiveness of our proposed method on three video grounding datasets. On ActivityNet-Captions especially, our method obtains the accuracy of 42.49\%, which significantly outperforms the state-of-the-art, \ie, 36.90\% by He \etal\cite{he2019read}.
	\end{itemize}

	\begin{figure*}[!t]
		\centering
		\includegraphics[width=\linewidth]{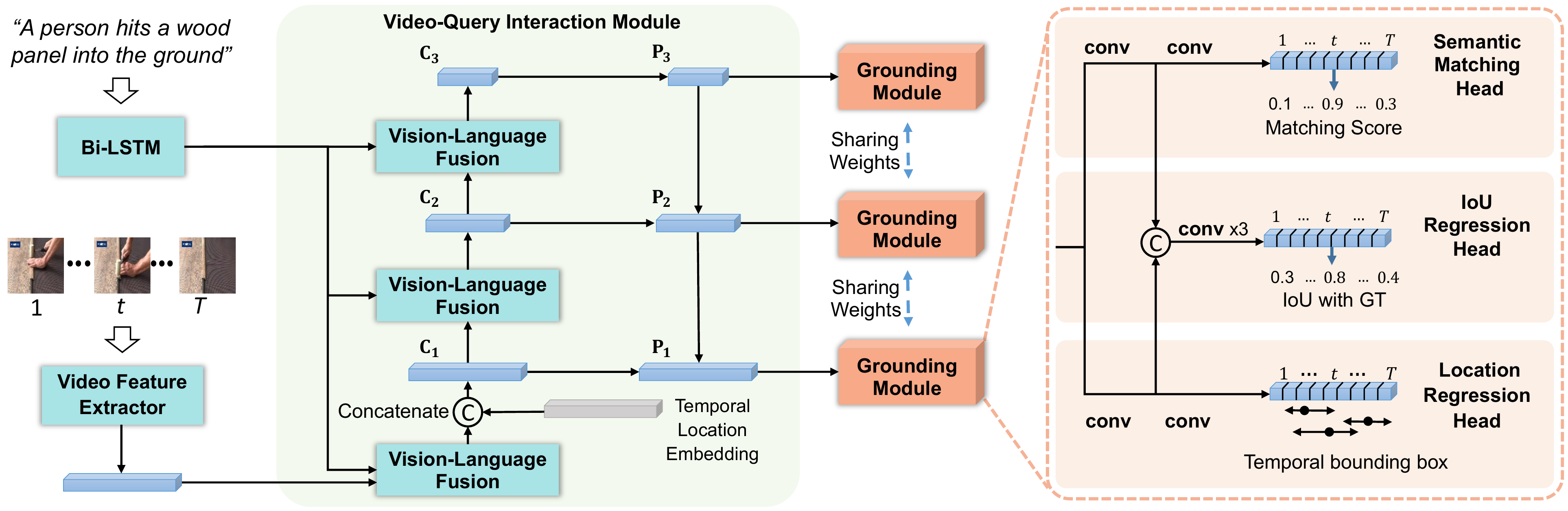}
		\caption{Schematic of our dense regression network. We use the video-query interaction module to fuse the features from the video and query. By constructing the feature pyramid, we obtain hierarchical feature maps and forward them to the grounding module. At each location $t$, the grounding module predicts a temporal bounding box, along with a semantic matching score and an IoU score for ranking.}
		\label{fig:overall}
	\end{figure*}
	
	\section{Related work}
	\noindent \textbf{Video grounding.}
	Recently, great progress has been achieved in deep learning~\cite{zhang2019collaborative,zhang2019whole,cao2019multi,guo2019nat,guo2020multi,guo2019auto,hu2019multi,pmlr-v80-cao18a,zhuang2018discrimination}, which facilitates the development of video grounding. Existing methods on this task can be grouped into two categories (\ie, two-stage and one-stage).
	Most two-stage methods~\cite{gao2017tall,hendricks2018localizing, ge2019mac,chen2018temporally,liu2018cross,zhang2019cross} resort to a propose-and-rank pipeline, where they first generate proposals and then rank them relying on the similarity between proposal and query. Gao~\etal~\cite{gao2017tall} and Hendricks~\etal~\cite{hendricks2018localizing} propose to use the sliding windows as proposals and then perform a comparison between each proposal and the input query in a joint multi-modal embedding space. To improve the quality of the proposals, Xu~\etal~\cite{xu2019multilevel} incorporate a query into a neural network to generate the query-guided proposals. Zhang~\etal~\cite{zhang2019man} explicitly model temporal relations among proposals using a graph. The two-stage methods are straightforward but have two limitations: 1) Comparing all the proposal-query pairs
	leads to a heavy computation cost; 2) The performance highly relies on the quality of proposals. Our method is able to avoid the above limitations since the candidate proposals are not required.

	To perform video grounding more efficiently, many methods that go beyond the propose-and-rank pipeline have been studied. He~\etal~\cite{he2019read} and  Wang~\etal~\cite{wang2019language} propose a reinforcement learning method for video grounding task. In the work by He~\etal~\cite{he2019read}, the agent adjusts the boundary of a temporal sliding window according to the learned policy. At the same time, Yuan~\etal~\cite{yuan2019to} propose the attention-based grounding approach which directly predicts the temporal coordinates of the video segment that described by the input query. Ghosh~\etal~\cite{ghosh2019excl} and Chen~\etal~\cite{chen2018temporally} propose to select the starting and ending frames by leveraging cross-modal interactions between text and video. Specifically, they predict two probabilities at each frame, which indicate whether this frame is a starting (or ending) frame of the ground truth video segment. Unlike the previous work by Chen ~\etal\cite{chen2018temporally} and Ghosh~\etal\cite{ghosh2019excl} where only the starting and ending frame are selected as positive training samples, our method is able to leverage much more positive training samples, which significantly boosts the grounding performance.

	\noindent \textbf{Anchor-free object detection.}
	Anchor-free object detectors~\cite{huang2015densebox,redmon2016you,tian2019fcos,law2018cornernet,kong2019foveabox} predict bounding boxes and class scores without using predefined anchor boxes. Redmon~\etal propose YOLOv1~\cite{redmon2016you} to predict bounding boxes at the points near the center of objects. Law~\etal propose CornerNet~\cite{law2018cornernet} to detect an object bounding box as a pair of corners and CornerNet obtains a high recall. Kong~\etal propose FoveaBox~\cite{kong2019foveabox} to predict category-sensitive semantic maps for the object existing possibility and produce a category-agnostic bounding box at each position. 
	Tian~\etal devise FCOS~\cite{tian2019fcos} to make full use of the pixels in a ground truth bounding box to train the model and propose “centerness” to suppress the low-quality predictions. Our work is related to FCOS since we also directly predict the distance from each frame to the ground truth boundary.

	\section{Proposed method}
	
	\noindent \textbf{Notation.}
	Let $V=\{I_{t} \in \mathbb{R}^{H\times W\times 3}\}_{t=1}^T$ be an untrimmed video, where $I_{t}$ denotes the frame at time slot $t$ with height $H$ and width $W$.
	We denote the query with $N$ words as $Q=\{w_n\}_{n=1}^{N}$, where $w_n$ is the $n$-th word in the query. 
	
	\noindent \textbf{Problem Definition.}
	Given a video $V$ and a query $Q$, video grounding requires the machine to localize a video segment (\ie, a temporal bounding box $\mathbf{b}=(t_s, t_e)$) starting at $t_s$ and ending at $t_e$, which corresponds to the query.
    This task is very challenging since it is difficult to localize actions of interest precisely in a video with complex contents. More critically, only a few frames are annotated in one video, making the training samples extremely imbalanced.
	
	\subsection{General scheme}
	\label{sec:general}
	
	
	
	We focus on solving the problem that existing video grounding methods neglect the rich information from the frames within the ground truth, which, however, is able to significantly improve the localization accuracy. 
	To this end, we propose a dense regression network to regress the starting (or ending) frame of the video segment described by the query for each frame. In this way, we are able to select every frame within the ground-truth as a positive training sample, which significantly benefits the training of our video grounding model.
	

	Formally, we forward the video frames $\{I_{t}\}_{t=1}^T$ and the query $\{w_n\}_{n=1}^{N}$ to the video-query interaction module $G$ for extracting the multi-scale feature maps. Then, each  feature map is processed by the grounding module, which consists of three components, \ie, location regression head $M_{loc}$, semantic matching head $M_{match}$ and IoU regression head $M_{iou}$. The \textbf{location regression head} predicts a temporal bounding box $\hat{\mathbf{b}}_{t}$ at the $t$-th frame by computing
	\begin{equation}\label{eq:general}
	\begin{split}
	\{\hat{\mathbf{b}}_{t}\}_{t=1}^T&=\{(t - \hat{d}_{t,s}, t + \hat{d}_{t,e})\}_{t=1}^T, \\
	\{(\hat{d}_{t,s}, \hat{d}_{t,e})\}_{t=1}^{T} 
	& = M_{loc} (G(\{I_{t}\}_{t=1}^T, \{w_n\}_{n=1}^{N})),
	\end{split}
	\end{equation}
	where $(\hat{d}_{t,s}, \hat{d}_{t,e})$ are the predicted distances to the starting and ending frame. With the predicted boxes $\{\hat{\mathbf{b}}_{t}\}_{t=1}^T$ at hand, our target is to select the box that matches the query best. To this end, we propose two heads in the grounding module.
	The \textbf{semantic matching head} predicts a score $\hat{m}_t$ indicating whether the content in the box $\hat{\mathbf{b}}_{t}$ matches the query semantically. However, this score cannot directly reflect the localization quality (\ie, the IoU with the ground truth), which, however, is also very important for video grounding. Therefore, we propose the \textbf{IoU regression head} to predict a score $\hat{u}_t$ for directly estimating the IoU between $\hat{\mathbf{b}}_{t}$ and the corresponding ground truth. 
	The schematic of our approach is shown in Figure~\ref{fig:overall}. For simplicity, we denote our model as \textbf{d}ense \textbf{r}egression \textbf{n}etwork (DRN).

	\textbf{Inference details.} Given an input video, we 
	forward it through the network and obtain a box $\hat{\mathbf{b}}_{t}$, a semantic matching score $\hat{m}_t$ as well as an IoU score $\hat{u}_t$ for each frame $I_{t}$. The final grounding result is obtained by choosing the box with the highest $\hat{m}_t \times \hat{u}_t$.
	
	In the following, we will introduce the details of the video-query interaction module in Section~\ref{sec:feature}. Then, we
	detail the location regression head, the semantic matching head  and the IoU regression head in Sections~\ref{sec:regression-head},  \ref{sec:matching-head}, and \ref{sec:iou-head}, respectively. Last, we introduce the training details of our model in Section~\ref{sec:training}. 
	
		\subsection{Multi-level video-query interaction module}
	\label{sec:feature}
	
	Building connections between vision and language is a crucial step for video grounding. To learn better vision-language representations, we propose a  multi-level video-query interaction module.
	Given a video with $T$ frames, we use some feature extractor (e.g., C3D~\cite{tran2015learning}) to obtain the video feature $\mathbf{F} \in \mathbb{R}^{T \times c}$, where $c$ is the channel dimension.  Then, the vision-language representations are produced by using multi-level fusion and temporal location embedding.
	
	\noindent \textbf{Multi-level fusion.}  The target video segments described by the query often have large scale variance in video grounding. For example on Charades-STA dataset~\cite{gao2017tall}, the shortest ground truth is 2.4s while the longest is 180.8s. To handle this issue, we follow Lin~\etal~\cite{lin2017feature} to obtain a set of hierarchical feature maps from multiple levels.
	Since the model may focus on different parts of the input query at each level, we follow~\cite{hu2018explainable} to fuse the query and the video features at different levels.
	Specifically, we encode the query $Q=\{w_n\}_{n=1}^{N}$ into $\{\Vec{h}_n\}_{n=1}^N$ and a global representation $g$ by using a bi-directional LSTM as:
	\begin{equation}
	\Vec{h}_1,\Vec{h}_2,\dots,\Vec{h}_{N} = \mathrm{BiLSTM}(Q)\ \text{and}\ g=[\Vec{h}_1;\Vec{h}_N],
	\end{equation}
	where $\Vec{h}_n = [\stackrel{\rightarrow}{h_n};\stackrel{\leftarrow}{h_n}]$
	is the concatenation of the forward and backward hidden states of the LSTM for the $n$-th word. For the $i$-th level, a textual attention $\alpha_{i,n}$ is computed over the words, and the query feature $\Vec{q}_i$ is computed as:
	\begin{equation}
	\begin{split}
	\Vec{q}_i & = \sum_{n=1}^{N} \alpha_{i,n} \cdot \Vec{h}_n,\\
	\alpha_{i,n} & = \mathrm{Softmax}(W_1(\Vec{h}_n \odot (W_2^{(i)} \text{ReLU}(W_{3}g)))), 
	\end{split}	
	\end{equation}
	where $\odot$ is element-wise multiplication. $W_1$ and $W_3$ are the parameters shared across different levels but $W_2^{(i)}$ is learned separately for each level $i$. Given the input visual feature $\mathbf{M}_i \in \mathbb{R}^{T_i \times c}$ of a vision-language fusion module, we first duplicate $\Vec{q}_i$ for $T_{i}$ times to obtain a feature map $\mathbf{D}_i \in \mathbb{R}^{T_i \times c}$, where $T_{i}$ is the temporal resolution at the $i$-th level. Then, we perform element-wise multiplication to fuse $\mathbf{M}_i$ and $\mathbf{D}_i$, leading to a set of feature maps $\{\mathbf{C}_i \in \mathbb{R}^{T_i \times c}\}_{i=1}^{L}$, where L is set to 3 in our paper. Last, we obtain the feature maps $\{\mathbf{P}_i \in \mathbb{R}^{T_i \times c}\}_{i=1}^{L}$ for the grounding module by using FPN. We put more details in the supplementary material.

	
	\noindent \textbf{Temporal location embedding.} We find that the queries often contain some words for referring temporal orders, such as ``after" and ``before". Therefore, we seek to fuse the temporal information of the video with the visual features.
	The temporal location of the $t$-th frame (or segment) is $\Vec{l}_t = [\frac{t-0.5}{T}, \frac{t+0.5}{T}, \frac{1}{T}]$. The location embedding $\Vec{l}_t$ is concatenated with the output of the vision-query fusion module that fuses the video feature $\mathbf{F}$ and the query feature. Note that the concatenation is performed along the channel dimension, resulting in the feature map $\mathbf{C}_1$. 
	
	


	\subsection{Location regression head}
	\label{sec:regression-head}
	
	With the vision-language representation $\mathbf{P}$ (we omit  index $i$ for better readability), we propose a location regression head to predict the distance from each frame to the starting (or ending) frame of the video segment that corresponds to the query.
	We implement it as two 1D convolution layers with two output channels in the last layer. 
	For each location $t$ on the feature map $\mathbf{P}$, if it falls inside the ground truth, then this location is considered as a training sample. Then, we have a vector $\Vec{d}_t=(d_{t,s}, d_{t,e})$ being the regression target at location $t$. Here, $d_{t,s}$ and $d_{t,e}$ denote the distance from location $t$ to the corresponding boundary and are computed as 
	\begin{equation}\label{eq:regression_target}
	d_{t,s} = t - t_s, d_{t,e} = t_e - t,
	\end{equation}
	where $t_s$ and $t_e$ is the starting and ending frames of the ground truth, respectively. 
	It is worth noting that $d_{t,s}$ and $d_{t,e}$ are all positive real values since the positive location $t$ falls in the ground truth (\ie, $t_s < t < t_e$). For those locations fall outside the ground truth, we do not use them to train the location regression head as in~\cite{tian2019fcos}. 
	
	It is worth mentioning that the FPN~\cite{lin2017feature} exploited in our video-query interaction module could also help the location regression head.
	The intuition is that all the positive locations from different feature maps can be used to train the location regression head, which further increases the number of training samples.

	\subsection{Semantic matching head}
	\label{sec:matching-head}
	
	For each video-query pair, the location regression head predicts a temporal bounding box $\hat{\mathbf{b}}_{t}$ at each location $t$. Then, how to select the box that matches the query best is the key to perform video grounding.
	
	Since the target of video grounding is to localize the video segments described by the query, it is straightforward to evaluate whether the content in $\hat{\mathbf{b}}_{t}$ matches the query semantically.
	To this end, we devise a semantic matching head to predict a score $\hat{m}_t$ for each predicted box $\hat{\mathbf{b}}_{t}$.
	The semantic matching head is implemented as two 1D convolution layers with one output channel in the last layer. 
	If location $t$ falls in the ground truth, its label is set as $m_t=1$.
	For those locations fall outside the ground truth, we consider them as negative training samples, \ie, $m_t=0$.


	\subsection{IoU regression head}
	\label{sec:iou-head}
	
	The semantic matching score $\hat{m}_t$ indicates whether the content in the box $\hat{\mathbf{b}}_{t}$ matches the query semantically. However, we also care about whether $\hat{\mathbf{b}}_{t}$ matches the ground truth temporal boundary, which can be measured by the localization quality (\ie, the IoU with the ground truth).
	
	To find the box with the best localization quality, one may use the ``centerness" technique in FCOS~\cite{tian2019fcos}. In short, ``centerness" is introduced for object detection to suppress the low-quality detected objects based on a hand-crafted assumption---the location closer to the center of objects will predict a box with higher localization quality (\ie, a larger IoU with the ground truth). However, we empirically found that this assumption is inapplicable to video grounding. Specifically, we conduct an experiment to find out which location predicts the best box (\ie, has the largest IoU with the ground truth).
	For each video-query pair, we select the predicted box that has the largest IoU with the ground truth.
	Then, we divide the ground truth into three portions evenly and sum up the number of locations that predicts the best box for each portion.
	Experimental results show that More than 46\% of the predictions are not predicted by the central locations of the ground truth.

	In this paper, we propose to explicitly consider the localization quality of the predicted box $\hat{\mathbf{b}}_{t}$ in the training and testing. The main idea is as straightforward as predicting a score at each location $t$ to estimate the IoU between $\hat{\mathbf{b}}_{t}$ and the corresponding ground truth. To do so, we train a three-layer convolution network as the IoU regression head in the grounding module, as shown in Figure~\ref{fig:overall}. Note that the input of the 
	IoU regression head is the concatenation of the feature maps obtained from the first convolution layer of the semantic matching head and the location regression head.
	The training target $u_t$ is obtained by calculating the IoU between $\hat{\mathbf{b}}_{t}$ and the corresponding ground truth. 
	


	\subsection{Training details}
	\label{sec:training}
	
	We define the training loss function for the location regression head as follows:
	\begin{equation}\label{eq:regression-loss}
	L_{loc}  =  \frac{1}{N_{pos}} \sum_{t=1}^{T} \mathbbm{1}_{gt}^t L_{1}(\Vec{d}_t, \hat{\Vec{d}}_t),
	\end{equation}
	where we use the IoU regression loss~\cite{yu2016unitbox} as  $L_{1}$ following Tian~\etal~\cite{tian2019fcos}. $N_{pos}$ is the number of positive samples. $\mathbbm{1}_{gt}^t$ is the indicator function, being 1 if location $t$ falls in the ground truth and 0 otherwise.
	The training loss function for the semantic matching head is defined as:
	\begin{equation}\label{eq:matching-loss}
	L_{match} = \frac{1}{N_{pos}} \sum_{t=1}^{T} L_{2}(m_t, \hat{m}_t),
	\end{equation}
	where we adopt the focal loss~\cite{lin2017focal} as $L_{2}$ since it is effective when handling the class imbalance issue. 
	To train the IoU regression head for predicting the IoU between the predicted box and ground truth, we define the training loss function as follows:
	\begin{equation}\label{eq:iou-loss}
	L_{iou} = \sum_{t=1}^T L_{3}(u_t, \hat{u}_t),
	\end{equation}
	where we choose to use the Smooth-L1 loss ~\cite{girshick2015fast} as $L_{3}$ because it is less sensitive to outliers.
	
%
%
%
%
	
	With randomly initialized parameters, the location regression head often fails to produce high-quality temporal bounding boxes for training the IoU regression head. Thus, we propose a three-step strategy to train the proposed DRN, which consists of a video-query interaction module $G$, a semantic matching head $M_{match}$, an IoU regression head $M_{iou}$ and a location regression head $M_{loc}$.
	Specifically, in the first step, we fix the parameters of the IoU regression head and train the DRN by minimizing Equations~\eqref{eq:regression-loss} and ~\eqref{eq:matching-loss}. In the second step, we fix the parameters in DRN except for the IoU regression head and train the DRN by minimizing Equation~\eqref{eq:iou-loss}. 
	In the third step, we fine-tune the whole model in an end-to-end manner\footnote{We put the training algorithm in the supplementary material.}.

	\section{Experiments}
	\subsection{Datasets}
	
	\noindent \textbf{Charades-STA} is a benchmark dataset for the video grounding task, which is built upon the Charades~\cite{sigurdsson2016hollywood} dataset. 
	The Charades dataset is collected for video action recognition and video captioning. 
    Gao~\etal~\cite{gao2017tall} adapt the Charades dataset to the video grounding task by collecting the query annotations. The Charades-STA dataset contains 6672 videos and involves 16128 video-query pairs, where 12408 pairs are used for training and 3720 for testing.  The duration of the videos is 29.76 seconds on average. Each video has 2.4 annotated moments and the duration of each moment is 8.2 seconds. We follow the same split of the dataset as in Gao~\etal~\cite{gao2017tall} for fair comparisons.
	
	\noindent \textbf{ActivityNet-Captions (ANet-Captions)} is collected for the dense video captioning task. It is also a popular benchmark for video grounding since the captions can be used as queries. ANet-Captions consists of 20K videos with 100K queries. 
	The videos are associated with 200 activity classes, where the content is more diverse compared to Charades-STA. 
	On average, each video contains 3.65 queries, and each query has an average length of 13.48 words. The average duration of the videos is around 2 minutes. The ActivityNet Captions dataset is split into the training set, validation set, testing set with a 2:1:1 ratio, including 37421, 17505 and 17031 video-query pairs separately. The public split of the dataset contains a training set and two validation sets val\_1 and val\_2. The testing set is withheld for competition. We train our model on the training set and evaluate it on val\_1 and val\_2 separately for fair comparisons. 
	
	\noindent \textbf{TACoS} dataset is collected by Regneri~\etal~\cite{regneri2013grounding} for video grounding and dense video captioning tasks. It consists of 127 videos on cooking activities with an average length of 4.79 minutes. For the video grounding task, TACoS dataset contains 18818 video-query pairs. 
	Compared to ActivityNet Captions dataset, TACoS has more temporally annotated video segments with queries per video. 
	Each video has 148 queries on average. Moreover, TACoS dataset is very challenging since the queries in TACoS dataset span over only a few seconds even a few frames. We follow the same split of the dataset as Gao~\etal~\cite{gao2017tall} for fair comparisons, which has 10146, 4589, and 4083 video-query pairs for training, validation, and testing respectively.
	
	\subsection{Implementation details}
	\label{sec:Implementation}
	
	\noindent \textbf{Evaluation metric.} 
    For fair comparisons, we follow Gao~\etal~\cite{gao2017tall} to compute ``R@$n$, IoU=$m$" as the evaluation metric. To be specific, it represents
    the percentage of testing samples that have at least one correct grounding prediction (\ie, the IoU between the prediction and the ground truth is larger than $m$) in the top-$n$ predictions.
	
	\noindent \textbf{Video Feature Extractor.} 
	We use the C3D~\cite{tran2015learning} network pre-trained on Sports-1M~\cite{karpathy2014large} as the feature extractor. The C3D network takes 16 frames as input and the outputs of the \textit{fc6} layer with dimensions of 4096 are used as a feature vector. We also extract the I3D~\cite{carreira2017quo} and VGG ~\cite{simonyan2014very} features to conduct experiments on Charades-STA. More details about the feature extractor are put in the supplementary material.  \\
	\noindent \textbf{Language Feature.}
	We transform each word of language sentences into lowercase. We use pre-trained GloVe word vectors to initialize word embeddings with the dimension of 300. A one-layer bi-directional LSTM with 512 hidden units serves as the query encoder. \\
	\noindent \textbf{Training settings.}
	The learning rate in the first training step is 0.001 and we decay it by a factor of 100 for the second step. During fine-tuning, we set the learning rate as $10^{-6}$. We set batch size as 32 and use Adam~\cite{kingma2014adam} as the optimizer.
	
   \begin{table}[!t]
	\caption{Comparisons with state-of-the-arts on Charades-STA.}
	\tabcolsep 3.5pt 
				\resizebox{1\linewidth}{!}{
	\begin{tabular}{l|c|c|c|c|c}
		\hline
		\multirow{2}{*}{Methods}             & \multirow{2}{*}{Feature} &      R@1       &      R@1       &      R@5       &      R@5       \\
		&                          &    IoU=0.5     &    IoU=0.7     &    IoU=0.5     &    IoU=0.7     \\ \hline
		CTRL~\cite{gao2017tall}                 &           C3D            &      23.63      &       8.89      &  58.92       &  29.52   \\
		SMRL~\cite{wang2019language}                 &           C3D            &      24.36      &      11.17      &       61.25     &  32.08  \\
		MAC~\cite{ge2019mac}                 &           C3D            &      30.48      &      12.20     &     64.84     &  35.13    \\
		T-to-C~\cite{xu2019multilevel} &           C3D            &      35.60      &      15.80      &  79.40        &   45.40      \\
		R-W-M~\cite{he2019read}                    &           C3D            &      36.70      &       -        &      -          &     -      \\
		DRN (ours)                                &           C3D            & \textbf{45.40} & \textbf{26.40} & \textbf{88.01}  & \textbf{55.38} \\ \hline\hline
		ExCL~\cite{ghosh2019excl}                                 &           I3D            &      44.10      &       22.40       &     -           &      -          \\
		DRN (ours)                                &           I3D            & \textbf{53.09}  & \textbf{31.75}  &  \textbf{89.06}     &   \textbf{60.05}          \\ \hline\hline
		SAP~\cite{chen2019semantic}                 &           VGG            &      27.42      &      13.36      &       66.37     &  38.15  \\
		MAN~\cite{zhang2019man}                                  &           VGG            &      41.24      &      20.54     &      83.21          &    51.85            \\
		DRN (ours)                                &           VGG            & \textbf{42.90} & \textbf{23.68} &\textbf{87.80}  & \textbf{54.87} \\ \hline
	\end{tabular}
	}
	\label{tab:charades}
\end{table}

\begin{table}[!t]
	\centering
	\caption{Comparisons on ANet-Captions using C3D features.
		($^{*}$) indicates the method that uses val\_2 split as the testing set, while other methods use the val\_1 split.}
		\resizebox{0.48\textwidth}{!}{
	\begin{tabular}{l|c|c|c|c}
		\hline
		\multirow{2}{*}{Methods}            &      R@1       &      R@1       &      R@5       &      R@5       \\
		&                IoU=0.5     &    IoU=0.7     &    IoU=0.5     &    IoU=0.7       \\ \hline
		CTRL~\cite{gao2017tall}                              &      14.00     &      -      &        -        &        -         \\
		ACRN~\cite{liu2018attentive}                    &      16.17     &      -      &        -        &        -         \\
		T-to-C\cite{xu2019multilevel}      &      27.70      &      13.60      &          59.20      &       38.30           \\
		R-W-M~\cite{he2019read} & 36.90       & -     & -  & -          \\ 
		DRN (ours)    &    \textbf{42.49}    & \textbf{22.25}   &  \textbf{71.85}  & \textbf{45.96}   \\ \hline\hline
	
		TGN$^{*}$~\cite{chen2018temporally}                   &     27.93      &       -      &        44.20        &         -        \\
		ABLR$^{*}$ \cite{yuan2019to}                     &      36.79      &       -        &      -          &        -         \\ 
		CMIN$^{*}$~\cite{zhang2019cross}  & 43.40  &   23.88  &  67.95  &\textbf{50.73}  \\   
		DRN$^{*}$ (ours)  & \textbf{45.45} & \textbf{24.36}   & \textbf{77.97} & 50.30  \\ \hline
	\end{tabular}
		}
	\label{tab:anet}
\end{table}

\begin{table}[!t]
	\centering
	\caption{Comparisons on TACoS using C3D features.}
	\resizebox{0.7\linewidth}{!}{
		\begin{tabular}{l|c|c}
			\hline
			\multirow{2}{*}{Methods} &     R@1             &      R@5           \\
			&                          IoU=0.5          &    IoU=0.5                              \\ \hline
			ABLR \cite{yuan2019to}             &     9.40           &    -            \\
			CTRL~\cite{gao2017tall}                               &     13.30           &         25.42        \\
			ACRN~\cite{liu2018attentive}             &     14.62          &    24.88           \\	
			SMRL~\cite{wang2019language}           &     15.95            &    27.84            \\
			CMIN~\cite{zhang2019cross}   & 18.05  &  27.02 \\
			TGN~\cite{chen2018temporally}                     &     18.90           &     31.02          \\
			MAC~\cite{ge2019mac}                           &     20.01           &    30.66   \\ \hline
			DRN (ours)                           & \textbf{23.17} &\textbf{33.36}       \\ \hline
	\end{tabular}}\label{tab:tacos}
 	\vspace{-0.1in}
\end{table}

	\subsection{Comparisons with state-of-the-arts}
	
	\noindent \textbf{Comparisons on Charades-STA.}
	We compare our model with the state-of-the-art methods in Table \ref{tab:charades}. Our DRN reaches the highest scores over all IoU thresholds. Particularly, when using the same C3D features, our DRN outperforms the previously best method (\ie, R-W-M~\cite{he2019read}) by 8.7\% absolute improvement, in terms of R@1, IoU=0.5. For fair comparisons with MAN~\cite{zhang2019man} and ExCL~\cite{ghosh2019excl}, we perform additional experiments by using the same features (\ie,  VGG and I3D) as reported in their papers. Our DRN outperforms them by 1.66\% and 8.99\%, respectively. 
	
	\noindent \textbf{Comparisons on ActivityNet-Captions.} Table \ref{tab:anet} reports the video grounding results of various methods. We follow the previous methods to use C3D features for fair comparisons. Since previous methods use different testing splits, we report the performance of our model on both val\_1 and val\_2.
	Regarding R@1, IoU=0.5, our method outperforms R-W-M~\cite{he2019read} by 5.59\% absolute improvement on val\_1 split and exceeds CMIN~\cite{zhang2019cross} by 2.05\% on val\_2 split.  
	
	\noindent \textbf{Comparisons on TACoS.} We compare our DRN with state-of-the-art methods with the same C3D features in Table~\ref{tab:tacos}. It is worth noting that this dataset is very challenging since each video may correspond to multiple queries (148 queries on average). Despite its difficulty, our method reaches the highest score in terms of both R@1 and R@5 when $IoU=0.5$ and outperforms previous best result by a large margin (\ie, 23.17\% vs. 20.01\%).

	\section{Ablation studies}
	\label{sec:ablation}
	
	In this section, we will perform complete and in-depth
	ablation studies to evaluate the effect of each component of our model. More details about the structures and training configurations of the baseline methods (such as DRN-Center) can be found in the supplementary material.
	
	\subsection{How does location regression help?}
	\label{sec:ablation_DRN}
	
	Compared with other one-stage video grounding methods, the key to our DRN is to leverage more positive samples for the training. Here, we implement three variants of our methods: \textbf{DRN-Half}, \textbf{DRN-Random} and \textbf{DRN-Center}. The three baselines are the same as the original DRN (\textbf{DRN-All}) except that they only select a subset of frames within the ground truth as the positive training samples. Specifically, DRN-Half randomly chooses 50\% of the frames within the ground truth to train the model. DRN-Random and DRN-Center are the extreme cases of our location regression settings, where they only randomly select one frame or the center frame within the ground truth as the positive training sample. By comparing the performance of the variants with our DRN, we justify the importance of increasing the number of positive training samples to train a one-stage video grounding model.
	Table~\ref{tab:sample_num} shows that all of these variants decrease the performance significantly. It verifies the effectiveness of our dense regression network, which is able to mine more positive training samples from sparse annotations.
	
	\begin{table}[!t]
		\centering
		\caption{Ablation study on the number of positive training samples on Charades, measured by R@1 and R@5 when IoU=0.5.}
		\begin{tabular}{l|c|c|c|c}
			\hline
			\multirow{2}{*}{Methods} &     R@1   &  \multirow{2}{*}{Gain}        &      R@5   &  \multirow{2}{*}{Gain}      \\
			&            IoU=0.5    &      &    IoU=0.5    &        \\ \hline
			DRN-Center              & 38.36       & -           & 83.36  & -   \\
			DRN-Random               & 40.88   & 2.52               & 84.11 & 0.75   \\
			DRN-Half              & 42.79     & 4.43             & 85.88   & 2.52   \\ 
			DRN-All &  \textbf{45.40} & \textbf{7.04} &   \textbf{88.01} & \textbf{4.65} \\ \hline
		\end{tabular}\label{tab:sample_num}
	\end{table}
	
	\subsection{Does IoU regression help video grounding?}
	
\begin{figure*}[!t]
	\includegraphics[width=\linewidth]{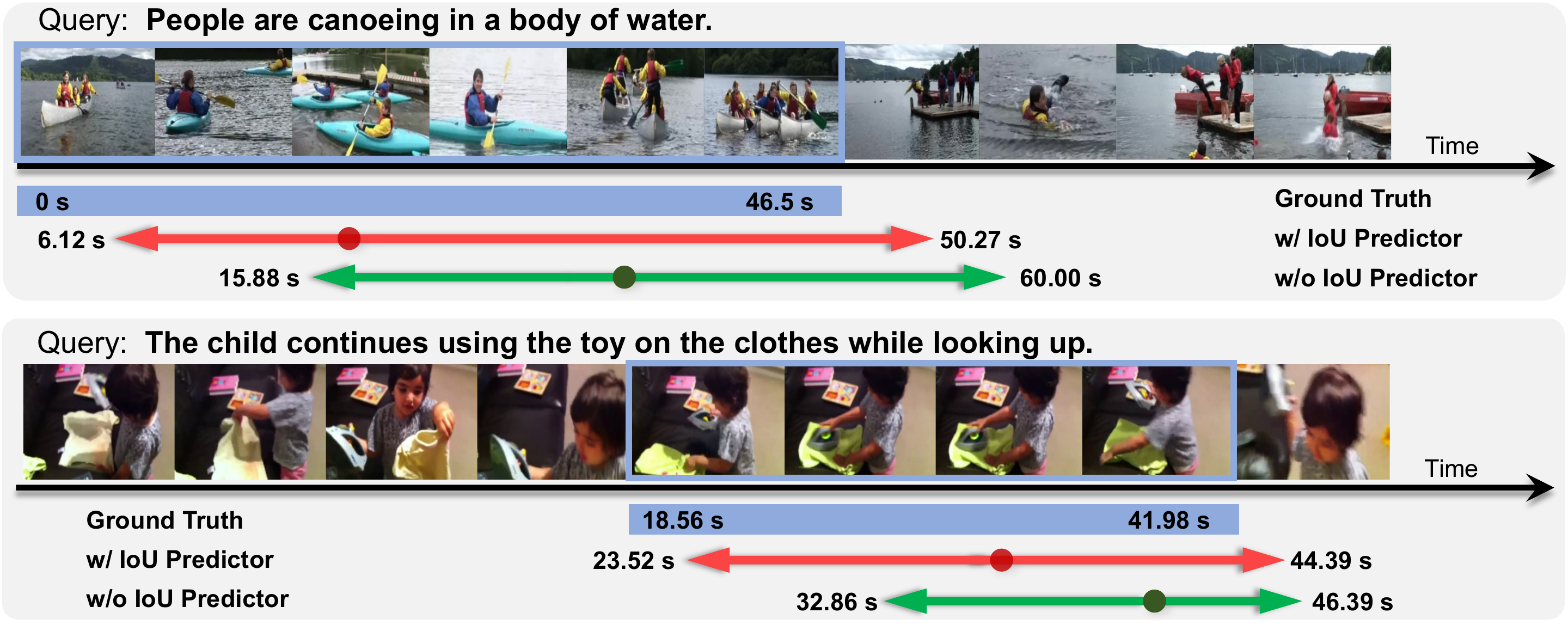}
		\caption{Qualitative results on ActivityNet Captions dataset.}
		\label{fig:visualization}
	\end{figure*}

	As discussed in Section {\ref{sec:iou-head}}, besides the IoU regression head, using ``centerness" technique is another way to assess the localization quality. Here, we implement a variant of our model by replacing the IoU regression head with the centerness head in FCOS~\cite{tian2019fcos}. Specifically, the centerness head is trained to predict a centerness score at each frame. The frame closer to the ground truth's center is expected to have a larger centerness value. In the inference, we follow \cite{tian2019fcos} to multiply the centerness score and matching score to obtain the final score for each predicted box.
	We also implement a baseline by removing the IoU regression head from our model and directly use the matching score to rank the predictions. 
	Table~\ref{tab:iou} reveals that the IoU regression head consistently improves the performance on both datasets. These results demonstrate that the matching score is not sufficient to evaluate the localization quality. Predicting the IoU between the predicted box and ground truth is straightforward and helpful for video grounding.
	Using centerness slightly
	decreases the grounding accuracy since the centerness assumption is not suitable for video grounding. We also visualize the qualitative results in Figure~\ref{fig:visualization}. In the top example, the two grounding results are both predicted by the frames within the ground truth, while the IoU regression head helps to select the one that has a larger IoU. In the bottom example, the background context is similar and the query is complex. Despite such difficulty, the IoU regression head still helps to select a better grounding result.
	More visualization results are shown in the supplementary material. 
	
	\begin{table}[!t]
		\centering
		\caption{Ablation study of the IoU regression head on Charades-STA and ActivityNet-Captions, measured by R@1 when IoU=0.5.}
		\begin{tabular}{l|l|c}
			\hline
			\multirow{2}{*}{Dataset} & \multirow{2}{*}{Methods} & R@1         \\
			&                          & IoU=0.5 \\ \hline
			\multirow{3}{*}{Charades-STA}     &    w/o IoU regression head      & 44.13  \\ 
			&    w/ Centerness          & 44.02    \\
			&    w/ IoU regression head     & \textbf{45.40}    \\
			\hline \hline
			\multirow{3}{*}{ANet-Captions}     &    w/o IoU regression head                   & 40.44       \\
			&    w/ Centerness                   & 39.83       \\
			&    w/ IoU regression head                   & \textbf{42.49}       \\
			\hline
		\end{tabular}\label{tab:iou}
	\end{table}

	\begin{table}[!t]
	\centering
	\caption{Ablation study of multi-level fusion (MLF) and location embedding on Charades-STA, measured by R@1 when IoU=0.5.}
	\begin{tabular}{l|cc|c}
		\hline
		\multirow{2}{*}{Dataset}       & \multicolumn{2}{c|}{Components} & R@1     \\ \cline{2-3}
		& MLF          & location         & IoU=0.5 \\ \hline
		\multirow{4}{*}{Charades-STA}  &    $\times$       &     $\times$  &  43.04       \\
		&    \checkmark         &       $\times$           &   43.79      \\
		&    $\times$          &    \checkmark            &   43.47      \\
		&    \checkmark         &     \checkmark    &  \textbf{45.40}       \\ \hline \hline
		\multirow{4}{*}{ANet-Captions} &    $\times$          &    $\times$              &     39.78    \\
		&    \checkmark          &     $\times$             &  40.61       \\
		&   $\times$           &    \checkmark              &  40.96       \\
		&     \checkmark         &     \checkmark             &  \textbf{42.49}     \\ \hline 
	\end{tabular}\label{tab:mlf}
\vspace{-0.1in}
\end{table}

	\subsection{Does multi-level fusion help?}
	
	The multi-level fusion (MLF) technique extracts different representations of the same query at different levels and fuses them with the video feature. Here, we implement a baseline by removing MLF from our DRN. Specifically, we only fuse the visual feature and the query feature at the first level (\ie, $C_1$ in Figure~\ref{fig:overall}). From Table~\ref{tab:mlf}, applying MLF to our model is able to lift the video grounding performance on both Charades-STA (43.79\% vs. 43.04\%) and ANet-Captions datasets (40.61\% vs. 39.78\%). In addition, we implement another baseline \textbf{MLF-Same} by using the same query feature to fuse the video feature at different levels. In our experiments, the \textbf{MLF-Same} baseline performs worse than our DRN on Charades-STA (44.76\% vs. 45.40\%), revealing that extracting different query features at different levels is able to improve the video-query representations and boost the grounding performance eventually. 
	
\begin{table}[!t]
	\centering
	\caption{Ablation study of the location embedding on the collected subset of ANet-Captions, measured by R@1 when IoU=0.5.}
	\begin{tabular}{l|l|l|c}
		\hline
		\multirow{2}{*}{Train} & \multirow{2}{*}{Test} & \multirow{2}{*}{Methods} & R@1         \\
		&      &                    & IoU=0.5 \\ \hline
		\multirow{2}{*}{Full-set}  & \multirow{2}{*}{Full-set}   &    w/o location      & 40.61   \\  
		&  &  w/ location         &  \textbf{42.49}   \\ \hline
		\multirow{2}{*}{Full-set}  & \multirow{2}{*}{Sub-set}   &    w/o location      & 47.38   \\ 
		&  &  w/ location         &  \textbf{48.37}   \\
		\hline 
		\multirow{2}{*}{Sub-set}  & \multirow{2}{*}{Sub-set}   &   w/o location                   & 43.28       \\
		&   & w/ location                 & \textbf{44.97 }     \\
		\hline
	\end{tabular}\label{tab:temporal}
			\vspace{-0.1in}
\end{table}

	\subsection{How does the location embedding help?}
	
	To evaluate the effectiveness of the temporal location embedding in our model, we conduct an ablation study by directly forwarding the video features into the network without concatenating with the location embeddings. The results in Table~\ref{tab:mlf} conclude that the location embedding makes the localization more precisely. One possible reason is that the model is able to learn the temporal orders of the video contents through the location embeddings.
	To further study the effect of the location embedding, we collect a ``temporal" subset of samples from the ANet-Captions dataset. In particular, we are interested in the query that contains four commonly used temporal words (\ie, before, after, while, then). The subset consists of 7176 training samples and 3620 testing samples. We use two settings to evaluate our model: 1) train on full ANet-Captions dataset and test on the temporal subset; 2) train and test on the temporal subset. From Table~\ref{tab:temporal}, using location embedding consistently improves the performance in both settings. Especially when training and testing the model on the temporal subset, the performance gain increases to 1.7\%, further verifying the effectiveness of the location embedding.
	
	
	\section{Conclusions}
	\label{sec:conclusions}
	
	In this paper, we have proposed a dense regression network for video grounding.
	By training the model to predict the distance from each frame to the starting (ending) frame of the video segment described by the query, the number of positive training samples is significantly increased, which boosts the performance of video grounding. Moreover, we have devised a simple but effective IoU regression head to explicitly consider the quality of localization results for video grounding.
	Our DRN outperforms the state-of-the-art methods on three benchmarks, \ie, Charades-STA, ActivithNet-Captions and TACoS. It would be interesting to extend our DRN for temporal action localization and dense video captioning, and we leave it for our future work.
	
	{\flushleft \bf Acknowledgements}. This work was partially supported by Guangdong Provincial Scientific and Technological Funds under Grant 2018B010107001, Grant 2019B010155002,
	National Natural Science Foundation of China (NSFC) 61836003 (key project), Program for Guangdong Introducing Innovative and Enterpreneurial Teams 2017ZT07X183,  
    Tencent AI Lab Rhino-Bird Focused Research Program (No. JR201902),
    Fundamental Research Funds for the Central Universities D2191240.


\newpage
\normalsize
\setcounter{section}{0}
\renewcommand\thesection{\Alph{section}}
\setcounter{figure}{0}
\renewcommand\thefigure{\Alph{figure}}
\setcounter{table}{0}
\renewcommand\thetable{\Alph{table}}

%

In the supplementary material, we first give the training details of our DRN in Section~\ref{sec:training}. Then, we
illustrate the details of the video-query interaction module in Section~\ref{sec:feature}. Next, we detail the grounding module in Section~\ref{sec:grounding}, followed by more qualitative results in Section~\ref{sec:visual}.
Last, we provide more details of ``centerness" in Section~\ref{sec:centerness}.  .

\section{More details about our DRN}
\label{sec:training}

The training details of our proposed DRN are shown in Algorithm~\ref{alg:train}. With randomly initialized parameters, the location regression head often fails to produce high-quality temporal bounding box for training the IoU regression head. Thus, we propose a three-step strategy to train the proposed DRN.
Specifically, in the first step, we fix the parameters of the IoU regression head and train the DRN by minimizing Equations (5) and (6). In the second step, we fix the parameters in DRN except for the IoU regression head and train the DRN by minimizing Equation(7). In the third step, we fine-tune the whole model in an end-to-end manner.

\section{Details of video-query interaction module}
\label{sec:feature}

\begin{figure*}[!t]
	\centering
	\includegraphics[width=\linewidth]{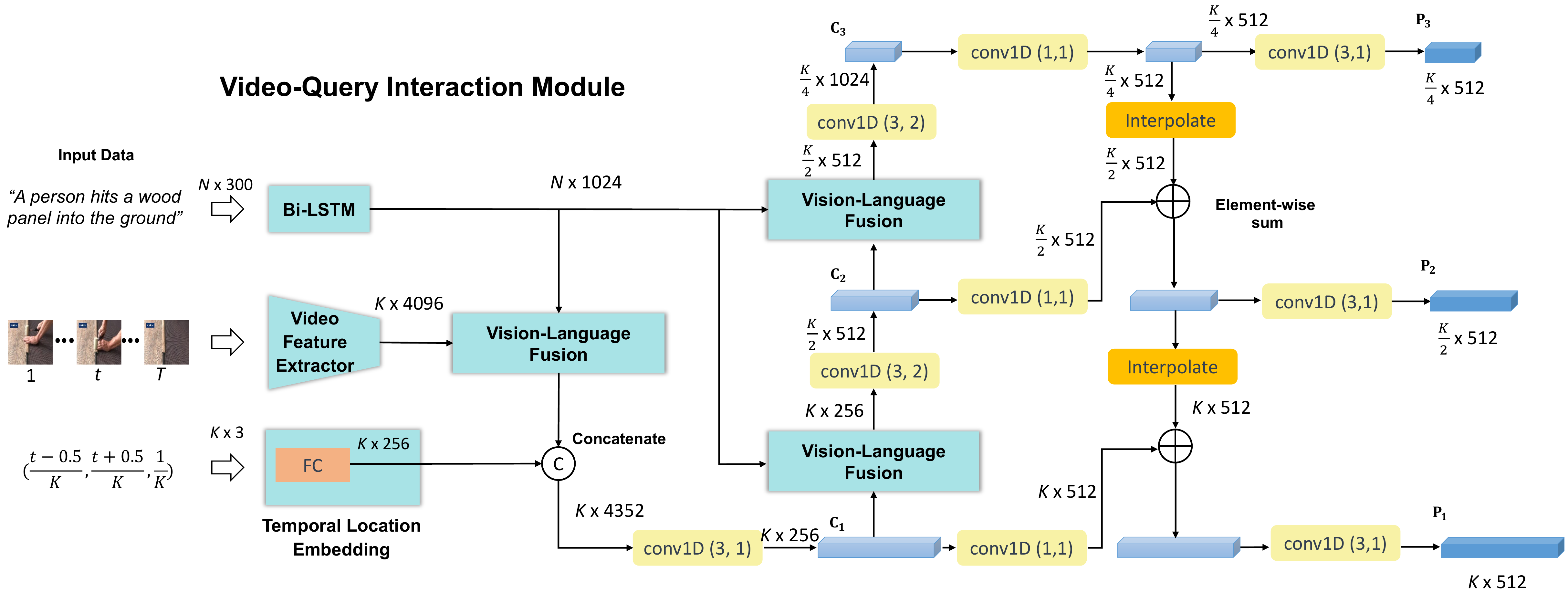}
	\caption{The details of Video-Query Interaction Module. Note that ``Conv1D ($b$, $s$)" denotes a 1D convolution layer with a kernel size of $b$ and a stride of $s$. All the convolution layers are followed by batch normalization and ReLU.}
	\label{fig:part2_interaction}

	\centering
	\includegraphics[width=\linewidth]{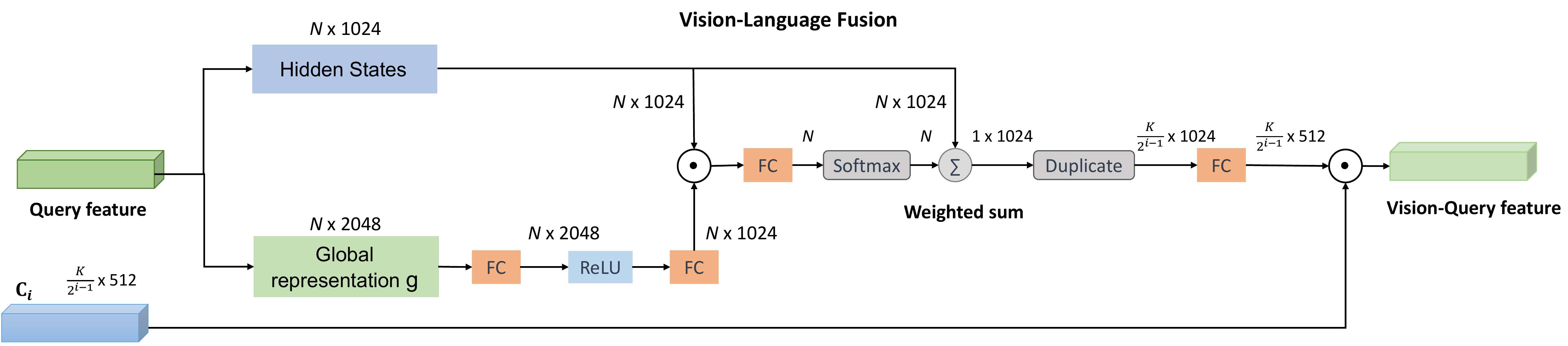}
	\caption{The details of Vision-Language-Fusion Module.}
	\label{fig:vision_language_fusion}
	\centering
	\includegraphics[width=\linewidth]{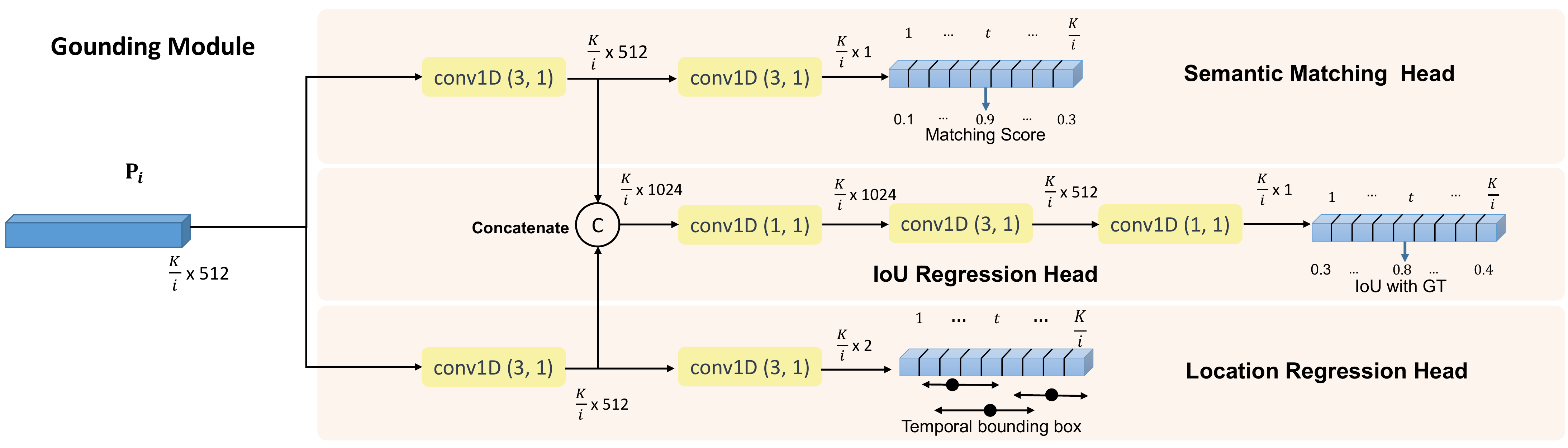}
	
	\caption{The details of  Grounding Module. The input $\mathbf{P}_{i}$ is from the $i$-th level in the feature pyramid with a temporal dimension of $\frac{K}{2^{i-1}}$.}
	\label{fig:ground_module}
\end{figure*}

The video-query interaction module consists of two parts, as shown in Figure~\ref{fig:part2_interaction}. The first part serves as a data preprocessor, which takes the query sentences, video frames and temporal coordinates as input and outputs the query feature and video feature ({C}$_{1}$). The second part is a fully convolutional network with vision-language fusion modules. It is 
used to fuse the video feature and query feature and construct a feature pyramid.
\begin{algorithm}[!t]
	\caption{Training details of DRN.}
	\textbf{Input:} Video $V=\{I_{t}\}_{t=1}^T$; query $Q=\{w_n\}_{n=1}^{N}$
	
	\textbf{Step1:} Fix the parameters of $M_{iou}$ 
	\begin{algorithmic}[1]
		\While {not converges}
		\State predict matching score $\hat{{m}}_t$
		\State predict regression offset $\hat{\Vec{d}}_t$ using Equation (1)
		\State update DRN by minimizing Equations (5) and (6)
		\EndWhile
	\end{algorithmic} 
	
	\textbf{Step2:} Fix the parameters of $G$, $M_{match}$, and $M_{loc}$ 
	\begin{algorithmic}[1]
		\While {not converges}
		\State predict bounding box $\hat{\mathbf{b}}_t$ using Equation (1)
		\State predict IoU between $\hat{\mathbf{b}}_{t}$ and ground truth
		\State update DRN by minimizing Equation (7)
		\EndWhile
	\end{algorithmic} 
	
	\textbf{Step3:} Fine-tune $G$, $M_{match}$, $M_{loc}$, and $M_{iou}$ jointly 
	\begin{algorithmic}[1]
		\While {not converges}
		\State predict matching score $\hat{m}_t$
		\State predict bounding box $\hat{\mathbf{b}}_{t}$ using Equation (1)
		\State predict IoU between $\hat{\mathbf{b}}_{t}$ and ground truth
		\State update DRN by minimizing Equations (5), (6), (7)
		\EndWhile
	\end{algorithmic}
	\textbf{Output:} Trained DRN
	
	\label{alg:train}
\end{algorithm}
\vspace{-0.1in}

\subsection{Video feature extractor}

Instead of predicting a temporal bounding box at each frame, we exploit a more efficient way to implement our dense regression network. Specifically, we divide a video into $K$ segments evenly. Thus, the temporal resolution of the video comes to $K$, which significantly reduces the computation in our model. 
Then, we use our model to predict a temporal bounding box \wrt the central frame of each segment.
We set $K$ as 32 for Charades-STA and ActivityNet Captions, and 128 for TACoS dataset. Three types of feature extractor are detailed as follows: \\
\noindent \textbf{C3D}. We use C3D [37] pre-trained on sport1M [22] to extract features. The C3D network takes 16 consecutive frames (a snippet) as input and the output of \textit{fc}6 layer is used as a snippet-level feature vector. The feature of each segment is obtained by performing max-pooling among the snippet-level features that correspond to the segment.\\
\noindent \textbf{VGG}.
We use VGG16-BN [34] pre-trained on ImageNet. VGG16-BN takes one frame as input and the output of \textit{fc}7 layer is used as the frame-level feature.
The segment feature is obtained by performing max-pooling among the frame-level features that correspond to the segment.\\
\noindent \textbf{I3D}.
We use I3D [3] pre-trained on Kinetics to extract features. The I3D network takes 64 consecutive frames (a snippet) as input and outputs a snippet-level feature vector. The feature of each segment is obtained by performing max-pooling among the snippet-level features that correspond to the segment.

\subsection{Query feature extractor}

First, each word in the input query sentence is mapped into a 300-dim vector using pre-trained GloVe word embeddings. Then, the word embeddings of the query sentence are fed into a one-layer bi-directional LSTM with 512 units. Last, the sequence of hidden states is used as query features. The hidden states of the first and the last word are concatenated, leading to the global representation $g$.


\subsection{Location embedding}

The input temporal coordinates of the $k$-th segment is a 3D vector, \ie, $(\frac{k-0.5}{K}, \frac{k+0.5}{K}, \frac{1}{K})$. We forward it to a linear layer, leading to a 256D location embedding. The location embedding is then concatenated with the video features along the channel dimension.

\subsection{Vision-Language Fusion Module}

We apply the textual attention mechanism to the input query feature and obtain the attended feature. Then, the attended query features and the features from a lower level of the pyramid are fused by using element-wise multiplication. The details are shown in Figure~\ref{fig:vision_language_fusion}.

\begin{figure*}[!t]
	\centering
	\includegraphics[width=.9\linewidth]{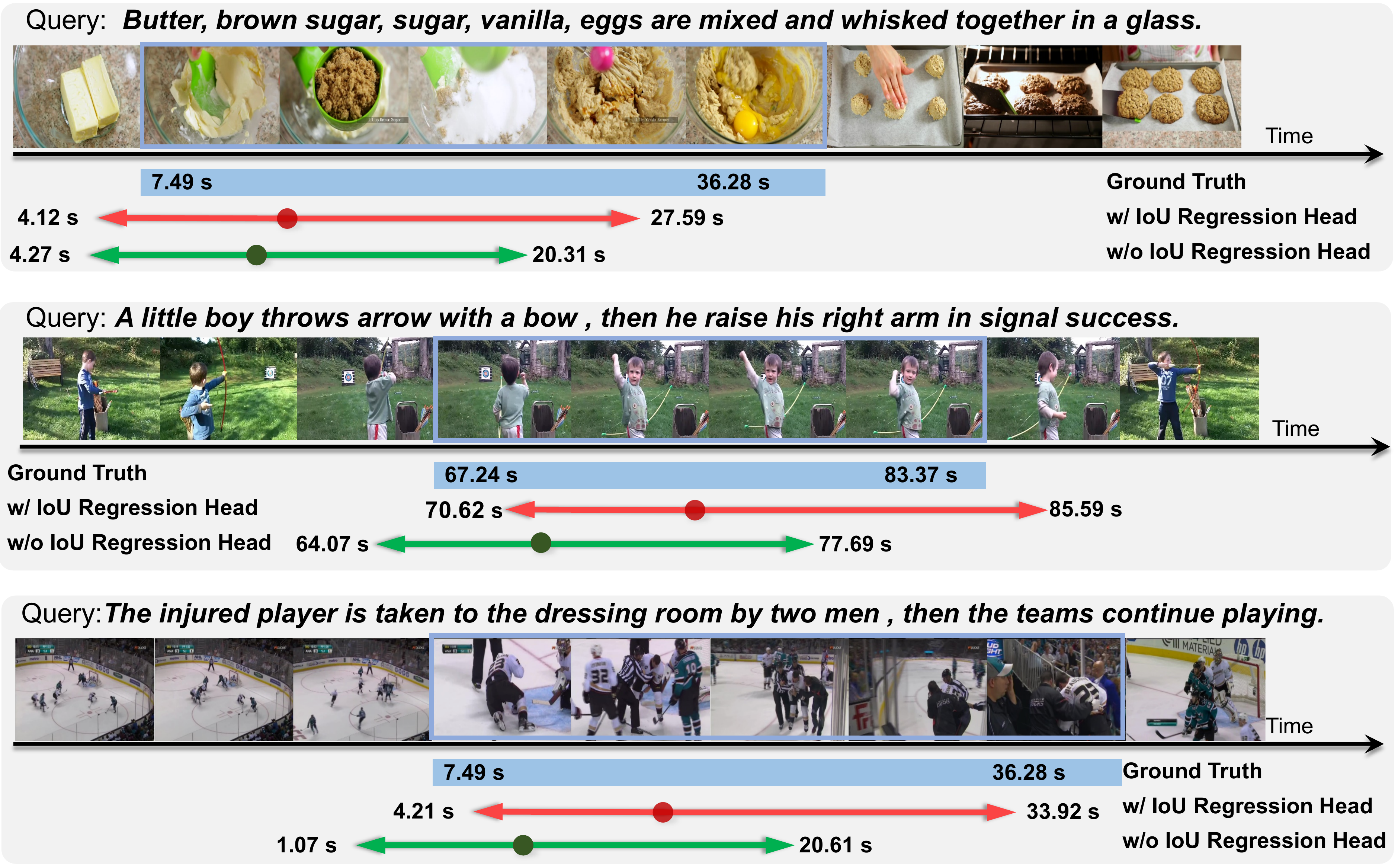}
	\centering
	\caption{Qualitative results.}
	\label{fig:visualizations}
\end{figure*}

\section{More details about grounding module}
\label{sec:grounding}

The grounding module involves three components, including semantic matching head, location regression head and IoU regression head.
Both of the semantic matching head and location regression head consist of two 1D convolution layers, and IoU regression head contains three 1D convolution layers. 
The details are shown in Figure~\ref{fig:ground_module}.

\section{More visualization examples}
\label{sec:visual}

We show more qualitative results of the IoU regression head in Figure~\ref{fig:visualizations}. The IoU regression head helps to select the prediction that has a larger IoU with the ground truth.

\section{More details about centerness}
\label{sec:centerness}

\subsection{Details of centerness baseline}

To compare our IoU regression with the centerness in FCOS [28], we conduct an experiment by replacing the loss function of IoU regression head with a centerness loss as in [28]. Specifically, we train the model to predict a centerness score for each location. The training target is defined as:
\begin{equation} 
centerness* = \sqrt{\frac{min(d_{t,s}^{*}, d_{t,e}^{*})}{max(d_{t,s}^{*}, d_{t,e}^{*})}}
\end{equation}
where $d_{t,s}^{*}$, $d_{t,e}^{*}$ are the distances between location $t$ and the starting frame, the ending frame of ground truth boundary respectively. We follow [28] to adopt the binary cross-entropy loss as the loss function for centerness in our experiments.


\subsection{Results of the centerness assumption}

\begin{figure}[!t]
	\centering
	\includegraphics[width=\linewidth]{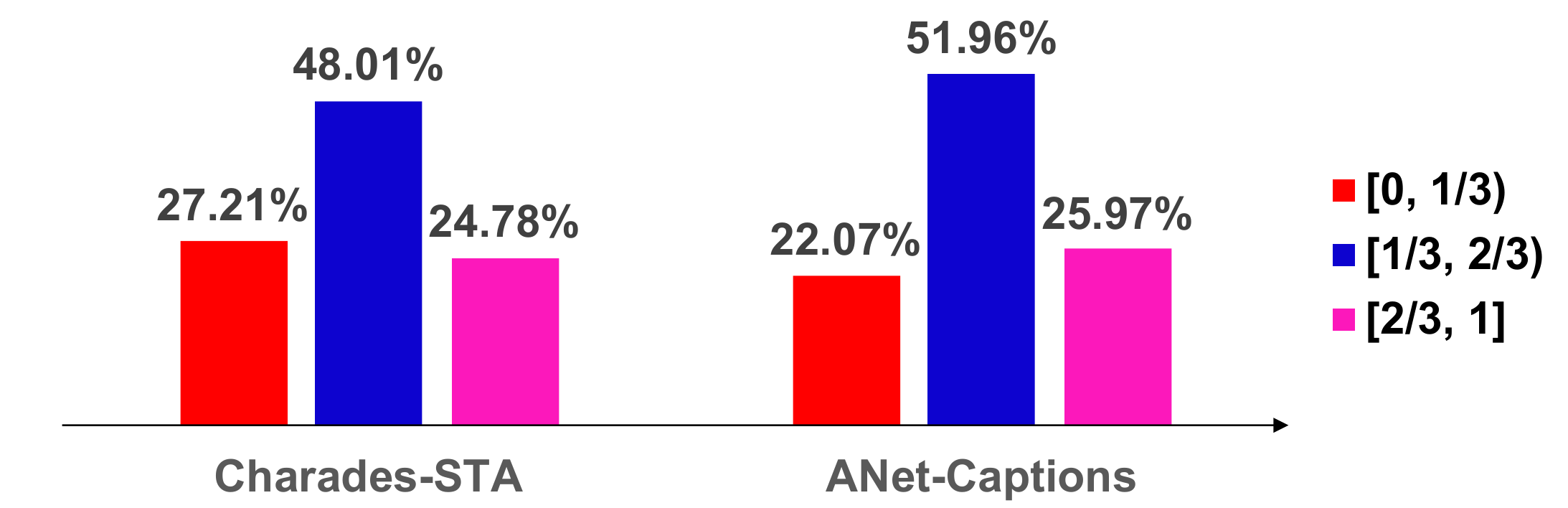}
	\caption{The location distribution of the ``best location" on two datasets. Here, the ``best location" denotes the location that predicts the best grounding result for each video-query pair. We show the statistics of their relative locations \wrt the ground truth, which has been divided into three portions evenly. Here, we only focus on the locations within the ground truth since few locations fall outside of the ground truth.}
	\label{fig:centerness}
\end{figure}

The centerness assumption [28] is that the location closer to the center of objects will predict a box with higher localization quality (\ie, a larger IoU with the ground truth). We conduct an experiment to find out which location predicts the best box. In our experiment, we train a model using the semantic matching loss and location regression loss.
For each video-query pair, we select the predicted box that has the largest IoU with the ground truth.
Then, we divide the ground truth into three portions evenly and sum up the number of the locations that predicts the best box for each portion.
From Figure~\ref{fig:centerness}, more than 48\% of the predictions are not predicted by the central locations of the ground truth.

\end{document}